\documentclass[conference]{IEEEtran}
\IEEEoverridecommandlockouts
\usepackage{cite}
\usepackage{amsmath,amssymb,amsfonts}
\usepackage{algorithmic}
\usepackage{graphicx}
\usepackage{textcomp}
\usepackage{xcolor}
\usepackage[colorlinks,linkcolor=red,anchorcolor=red,citecolor=blue]{hyperref}
\usepackage[lined,boxed,ruled]{algorithm2e}
\usepackage{paralist}
\graphicspath{{./figs/}}

\def\BibTeX{{\rm B\kern-.05em{\sc i\kern-.025em b}\kern-.08em
    T\kern-.1667em\lower.7ex\hbox{E}\kern-.125emX}}
\begin{document}

\title{
DarwinNet: An Evolutionary Network Architecture for Agent-Driven Protocol Synthesis}

\author{\IEEEauthorblockN{Jinliang Xu$^{*}$, Bingqi Li}
\IEEEauthorblockA{\textit{China Academy of Information and Communications Technology}, Beijing, China \\
xujinliang@caict.ac.cn}
}

\maketitle

\begin{abstract}
Traditional network architectures suffer from severe protocol ossification and structural fragility due to their reliance on static, human-defined rules that fail to adapt to the emergent edge cases and probabilistic reasoning of modern autonomous agents. To address these limitations, this paper proposes DarwinNet, a bio-inspired, self-evolving network architecture that transitions communication protocols from a \textit{design-time} static paradigm to a \textit{runtime} growth paradigm. DarwinNet utilizes a tri-layered framework—comprising an immutable physical anchor (L0), a WebAssembly-based fluid cortex (L1), and an LLM-driven Darwin cortex (L2)—to synthesize high-level business intents into executable bytecode through a dual-loop \textit{Intent-to-Bytecode} (I2B) mechanism. We introduce the Protocol Solidification Index (PSI) to quantify the evolutionary maturity of the system as it collapses from high-latency intelligent reasoning (Slow Thinking) toward near-native execution (Fast Thinking). Validated through a reliability growth framework based on the Crow-AMSAA model, experimental results demonstrate that DarwinNet achieves anti-fragility by treating environmental anomalies as catalysts for autonomous evolution. Our findings confirm that DarwinNet can effectively converge toward physical performance limits while ensuring endogenous security through zero-trust sandboxing, providing a viable path for the next generation of intelligent, self-optimizing networks.
\end{abstract}


\section{Introduction}
\label{sec:introduction}

For decades, the classical edifice of computer science has been built upon the aesthetics of completeness and determinism. Nowhere is this more evident than in the \textit{narrow waist} of the TCP/IP architecture. While this rigid design enabled the global expansion of the Internet, it has inadvertently led to a phenomenon known as \textit{protocol ossification} \cite{ammar2018ex, edeline2019bottom}. In modern heterogeneous environments—ranging from high-latency satellite constellations to massive-scale IoT deployments—the standardization cycle of traditional protocols is prohibitively long, often taking years to ratify minor adjustments to headers or congestion control logic.
Furthermore, the traditional paradigm of \textit{Code is Law} assumes that engineers can pre-emptively define every combination of input and state during the design phase. However, this reliance on \textit{mechanical completeness} creates inherently brittle systems. When faced with undefined edge cases or novel environmental stressors, rule-based systems either crash or fall into unknown error states, lacking the flexibility to adapt without human intervention.

As we enter the era of Large Language Models (LLMs) and autonomous Agents, the fundamental nature of a network node is changing \cite{wang2024survey, marro2024scalable}. Computing nodes are no longer merely passive executors of deterministic instructions; they have evolved into intelligent agents capable of probabilistic reasoning and intent understanding. In this context, legacy deterministic communication protocols have become a bottleneck, sacrificing efficiency for human readability—a constraint we term \textit{Carbon Hegemony}\footnote{Here Carbon Hegemony refers to the constraints imposed by human-centric cognitive structures on machine-to-machine communication, rather than environmental carbon emissions.}. DarwinNet seeks to dismantle this by envisioning a formal parting of ways between computer protocols and human language, allowing machines to communicate via high-dimensional logic that transcends human cognitive structures.
Crucially, this evolution does not imply the abandonment of existing network foundations. Instead, DarwinNet is positioned as a high-order augmentation for the 6G era, designed to handle hyper-heterogeneous edge scenarios where traditional manual standardization reaches its scalability limits, while still relying on the global reachability provided by the TCP/IP \textit{narrow waist} as its foundational anchor.

\begin{figure}[htbp]
\centerline{\includegraphics[width=\linewidth]{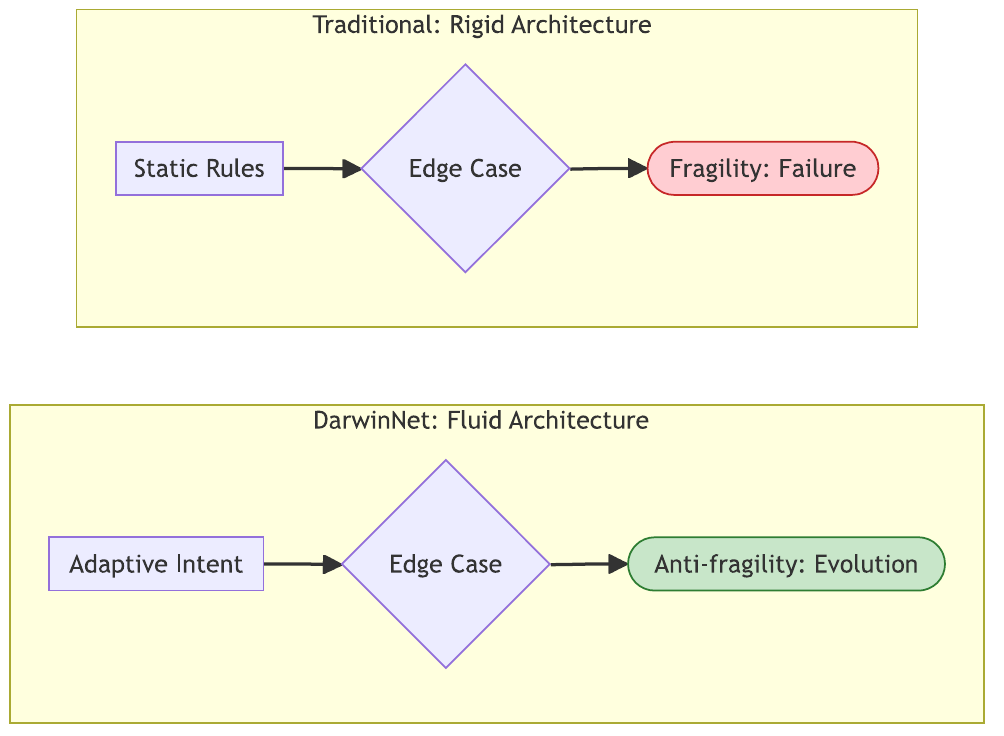}}
\caption{The Paradigm Shift: Contrasting the Fragility of Static Protocol Rules with the Anti-fragility of DarwinNet’s Fluid Evolution.}
\label{fig:comparison}
\end{figure}

To realize the transition from structural fragility to evolutionary anti-fragility depicted in Fig. \ref{fig:comparison}, this paper introduces \textbf{DarwinNet}---a bio-inspired, agent-driven network architecture that enables protocols to be \textit{grown} rather than \textit{designed.} Unlike traditional networks that succumb to ossification under static standards, DarwinNet conceptualizes the network as a living ecosystem capable of autonomous adaptation \cite{nezami2025connectivity}. Structurally, this is achieved through a tri-layered decoupling: a \textit{Physical/Mathematical Anchor} (Layer 0) for immutable constraints, a \textit{Fluid Cortex} (Layer 1) serving as the adaptive execution body, and a \textit{Darwin Cortex} (Layer 2) acting as the intelligent decision-making brain. By shifting the paradigm from rigid header-matching to high-dimensional intent negotiation, DarwinNet allows nodes to synthesize and deploy specialized communication logic in real-time. This mechanism ensures that network anomalies are no longer fatal errors, but rather catalysts for evolution that drive the system toward higher efficiency and survival.

In summary, DarwinNet represents a fundamental reconstruction of the network interaction model, moving from rigid rules to organic adaptation \cite{gao2025data}. The core contributions of this work are as follows:
\begin{inparaenum}
    \item \textbf{Paradigm Shift from Design to Growth:} We propose a self-adaptive network architecture that achieves \textit{anti-fragility}. By allowing protocols to evolve as fluid expressions of node intents, the network can autonomously adapt to environmental mutations—such as unforeseen congestion or novel attack patterns—without human-issued patches.  
    \item \textbf{Intent-to-Bytecode (I2B) Synthesis with Dual-Loop Mechanism:} We design an adaptive framework that leverages LLMs to transform high-level business intents into restricted, executable WebAssembly (WASM) bytecode. This is governed by a \textit{dual-loop} architecture: an \textit{outer loop} for cognitive protocol synthesis and an \textit{inner loop} for millisecond-level \textit{Hot Swaps} within a zero-trust sandbox \cite{liu2025webassembly}.  
    \item \textbf{Protocol Solidification Index (PSI):} We define a novel metric to quantify the maturity of the evolving system. This index captures the process of \textit{Protocol Liquefaction,} where high-level reasoning (System 2) gradually collapses into highly efficient, compiled execution paths (System 1) as common interaction patterns stabilize.
    \item \textbf{Reliability Verification Paradigm:} We introduce a mathematical verification framework based on the Crow-AMSAA (Duane) model. By treating protocol mutations as a reliability growth process, we provide a formal method to ensure that autonomous evolution remains within safety boundaries while continuously reducing failure rates over time.
\end{inparaenum}

The remainder of this paper is organized as follows. Section \ref{sec:relatedworks} reviews the related work. Section \ref{sec:preliminaries} lists preliminaries of the work and Section \ref{sec:framework} details the proposed DarwinNet framework. Section \ref{sec:experiments}  discusses experimental evaluations, and Section \ref{sec:conclusion} concludes the paper.

\section{Related Work}\label{sec:relatedworks}
The evolution of network flexibility has been driven by several key technological movements. This section situates DarwinNet within the landscape of programmable planes, intent-driven systems, emergent communication, and endogenous security.

\textbf{Programmable Data Planes and eBPF.} The advent of P4 (Programming Protocol-independent Packet Processors) and eBPF (extended Berkeley Packet Filter) has shifted the networking paradigm from fixed-function hardware to programmable targets \cite{bosshart2014p4}. These technologies allow developers to define custom packet-processing logic and execute it at near-native speeds within the kernel or on specialized switching chips \cite{bosshart2014p4, liu2025webassembly}. However, existing programmable data planes primarily function as execution environments that require human-defined logic at compile-time or deployment-time. While they provide the necessary \textit{body} for flexible networking, they lack an autonomous \textit{brain} capable of real-time reasoning and logic synthesis \cite{mai2021network}. DarwinNet builds upon these execution primitives but elevates them by introducing an agent-driven layer that dynamically generates and swaps protocol logic during runtime.

\textbf{Intent-Based Networking (IBN).} Traditional IBN systems aim to bridge the gap between high-level business goals and low-level network configurations \cite{leivadeas2022survey}. Most current IBN implementations are limited to \textit{parameter-level} adjustments, such as automatically tuning TCP window sizes or re-routing paths based on pre-defined policies \cite{mai2021network}. DarwinNet pushes the boundaries of IBN into the domain of \textit{logic-level} synthesis \cite{mittelmann2022automated,marro2024scalable}. Instead of mapping intents to a library of static, human-written scripts, DarwinNet utilizes a \textit{fluid cortex} to synthesize entirely new protocol behaviors in the form of executable bytecode. This transition from \textit{Intent-to-Policy} to  I2B enables a degree of architectural agility that conventional IBN cannot achieve \cite{leivadeas2022survey, chen2021evaluating}.

\textbf{Emergent Communication in Multi-Agent Reinforcement Learning (MARL).} Research in MARL has demonstrated that autonomous agents can \textit{invent} their own communication protocols to solve collaborative tasks, often resulting in high-dimensional vector-based languages that are unintelligible to humans \cite{liang2024generative,mordatch2018emergence, das2017learning,liberman2017balls}. While these emergent protocols achieve high efficiency, they are often \textit{black boxes} embedded within neural weights, making them difficult to verify, port, or debug. While maintaining safety via WASM, DarwinNet acknowledges that human experts are no longer the primary coders of protocol syntax. This marks a transition of computer science from a pure engineering discipline to an experimental science, where humans act as gardeners of an evolving ecosystem rather than micro-managers of bit-level headers. By using WASM as the synthesis target \cite{woodcock2009formal, mittelmann2022automated}, DarwinNet ensures that evolved protocols are explicit, portable, and subject to formal safety constraints within a zero-trust sandbox \cite{woodcock2009formal}.

\textbf{Mimic Defense and Endogenous Security.} The theory of Mimic Defense, characterized by Dynamic Heterogeneous Redundancy (DHR), suggests that system security can be enhanced by introducing uncertainty and dynamic transformations to thwart attackers. DarwinNet inherently realizes this security philosophy through the mechanism of protocol mutation \cite{axenie2024antifragility}. By constantly evolving the syntax and semantics of communication—rather than relying on a static, standardized \textit{instruction manual} like TCP/IP—DarwinNet creates a Moving Target Defense (MTD) at the protocol level. This dynamic nature significantly increases the cost of reconnaissance and exploitation for attackers, as the network's \textit{attack surface} is in a state of constant, intelligent flux.

\textbf{Recursive InterNetwork Architecture (RINA).} RINA represents a fundamental rethink of the OSI model, proposing a recursive structure that treats networking as a single type of layer (the Distributed IPC Facility) repeated at different scales \cite{grasa2022recursive, feldmann2007internet}. While RINA provides an elegant theoretical framework for cross-layer optimization and de-layering, it still assumes a relatively stable set of mechanisms and policies. DarwinNet shares RINA’s vision of breaking down rigid layer boundaries but focuses more on the \textit{biological} growth of functions rather than the structural recursion of the architecture \cite{weyns2023vision}. DarwinNet’s \textit{liquefaction} process treats protocol functions as atomic components that can be organically reassembled and evolved based on environmental fitness rather than predefined recursive templates.

\section{Preliminaries}
\label{sec:preliminaries}

This section establishes the theoretical and technical foundations of DarwinNet, integrating cognitive psychology, secure sandboxing, and reliability growth modeling.

\subsection{Dual-Process Theory in Network Intelligence}
The cognitive architecture of DarwinNet is inspired by the \textit{Dual-Process Theory} \cite{kahneman2011thinking}, which bifurcates human cognition into two distinct systems:
\begin{itemize}
    \item \textbf{System 1 (Fast Thinking):} Operates automatically and quickly with little or no effort. In our context, this corresponds to the solidified, high-performance execution of protocol bytecode.
    \item \textbf{System 2 (Slow Thinking):} Allocates attention to effortful mental activities, such as complex reasoning and logic synthesis. In DarwinNet, LLM-based agents perform System 2 reasoning to handle environmental anomalies and synthesize new protocols
\end{itemize}
The evolution of a protocol in DarwinNet is essentially the process of \textit{liquid} System 2 reasoning collapsing into \textit{solid} System 1 reflexes to achieve maximum efficiency  \cite{wei2022chain, nascimento2023self,booch2021thinking}.

\subsection{WebAssembly and Sandbox Isolation}
To ensure that autonomously synthesized logic does not compromise node security, we utilize WASM as the fluid execution target. WASM provides a memory-safe, hardware-independent, and near-native performance environment \cite{liu2025webassembly}. Its capability for millisecond-level instantiation and hot-swapping allows DarwinNet to perform logic mutations without interrupting active data streams. Furthermore, the use of restricted WASM bytecode facilitates formal verification of synthesized logic, ensuring it adheres to the \textit{Constitution} defined in Layer 0 \cite{woodcock2009formal}.

\subsection{Crow-AMSAA Reliability Growth Model}
The evolutionary maturity of DarwinNet is modeled using the \textit{Crow-AMSAA (Duane) Power Law process}. This model describes the reliability growth of a system undergoing iterative \textit{discovery and repair} cycles \cite{duane2007learning}. The failure (or protocol mismatch) intensity function is defined as:
\begin{equation}\label{eq:fail}
\lambda(t) = \alpha \beta t^{\beta-1}
\end{equation}
where $\alpha$ is a scale parameter and $\beta$ is the shape (growth) parameter. In an evolving system like DarwinNet, a shape parameter $\beta < 1$ signifies that the system is learning from environmental stressors, leading to a decreasing rate of protocol failures over time \cite{tang2015comparison, huang2005performance}. This provides the mathematical basis for our Protocol  PSI.

\section{The Proposed Framework}
\label{sec:framework}

This section delineates the structural and functional design of DarwinNet. Drawing inspiration from biological evolution and cognitive psychology, DarwinNet replaces the static OSI stack with a tri-layer architecture that separates immutable constraints from fluid, evolving logic.

\begin{figure*}[htbp]
\centerline{\includegraphics[width=7in]{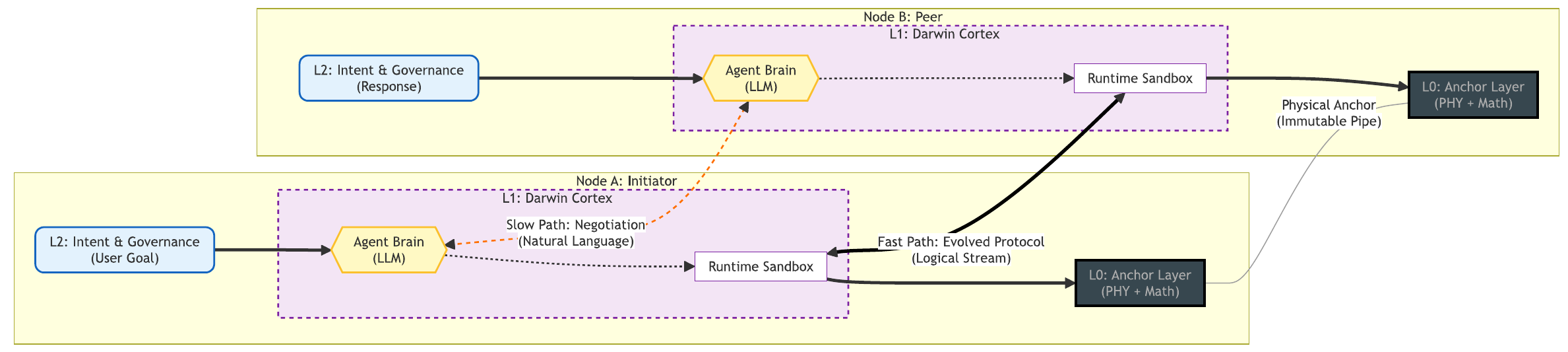}}
\caption{The DarwinNet Inter-node Interaction Framework: Illustrating the vertical functional decoupling from Intent (L2) to Anchor (L0) and the horizontal bifurcation of the Slow Negotiation Path and the Fast Execution Path.}
\label{fig:layered_framework}
\end{figure*}

Fig. \ref{fig:layered_framework} provides a global architectural overview of DarwinNet, depicting the interaction between two peer nodes (Node A and Node B). The architecture is characterized by a clear vertical hierarchy and distinct horizontal communication channels:
\begin{inparaenum}
    \item \textbf{Vertical Functional Flow:} Within each node, user goals are captured at the \textit{Intent Layer (L2)}, which serves as the governance center. These intents are passed down to the \textit{Darwin Cortex (L1)}, where an LLM-based Agent Brain synthesizes specific protocol logic. This logic is subsequently deployed into a \textit{Runtime Sandbox} (e.g., WASM) for execution. Finally, the \textit{Anchor Layer (L0)} provides the immutable foundation, ensuring physical connectivity and mathematical integrity. 
    \item \textbf{Horizontal Interaction Paths:} The system operates via three parallel interaction planes. First, the \textit{Slow Path} (dashed line) facilitates high-level negotiation between Agent Brains using natural language or semantic vectors. This path is responsible for protocol discovery and mutation. Second, the \textit{Fast Path} (solid bold line) represents the optimized data stream flowing between sandboxed execution environments, utilizing the evolved protocol logic for peak efficiency. Third, the \textit{Physical Anchor} (bottom line) signifies the underlying, unchangeable pipe (PHY/MAC) that supports the entire structure.
\end{inparaenum}

This tri-layered, dual-path design ensures that DarwinNet can decouple the \textit{intelligence of the protocol} from the \textit{mechanics of the transmission,} allowing nodes to evolve their communication language without disrupting the underlying physical connectivity.

\subsection{Layered Architecture: From Anchor to Fluid}
DarwinNet is organized into three distinct layers, transitioning from \textit{Physical/Mathematical Hardness} at the base to \textit{Intelligent Liquidity} at the apex. 

\subsubsection{Layer 0: The Immutable Anchor (The Constitution).} 
Layer 0 represents the fundamental \textit{laws of physics} of the network that remain constant throughout the evolutionary process. This layer ensures basic bit-level connectivity (e.g., the standard TCP/IP stack for physical reachability) and enforces mathematical truths, such as cryptographic verification and logical consistency ($1+1=2$). In this paradigm, we relinquish micro-control over the communication process (the Silicon-driven protocol) and retain only mathematical constraints over the results (the Consensus). Layer 0 ensures that mathematics remains the only universal language between carbon-based intent and silicon-based execution, serving as the final tether of human control over autonomous evolution \cite{feldmann2007internet}.

This layer functions as the \textit{Mathematical Constitution} of the network. No matter how complex or liquid the evolved logic in Layer 1 becomes, it is strictly forbidden from violating the physical and security predicates pre-defined by human architects in Layer 0. This ensures that the autonomy of the network is always bounded by a \textit{Human-in-the-loop} safety framework, preventing any unpredictable or runaway protocol behavior.

\subsubsection{Layer 1: The Fluid Cortex (The Polymorphic Body).}

Layer 1 is the execution environment of the network, characterized by its \textit{liquid} state. Unlike traditional fixed-protocol implementations, the Fluid Cortex utilizes a high-performance, zero-trust sandbox (e.g., WebAssembly) to run protocol logic. This layer serves as the \textit{system's body} or the \textit{Fast Path} (System 1), executing synthesized bytecode at near-native speeds. It supports millisecond-level hot-swapping of logic—such as switching from a JSON-based parser to a specialized binary delta-compression algorithm—without interrupting data flows.

\begin{figure*}[htbp]
\centerline{\includegraphics[width=7in]{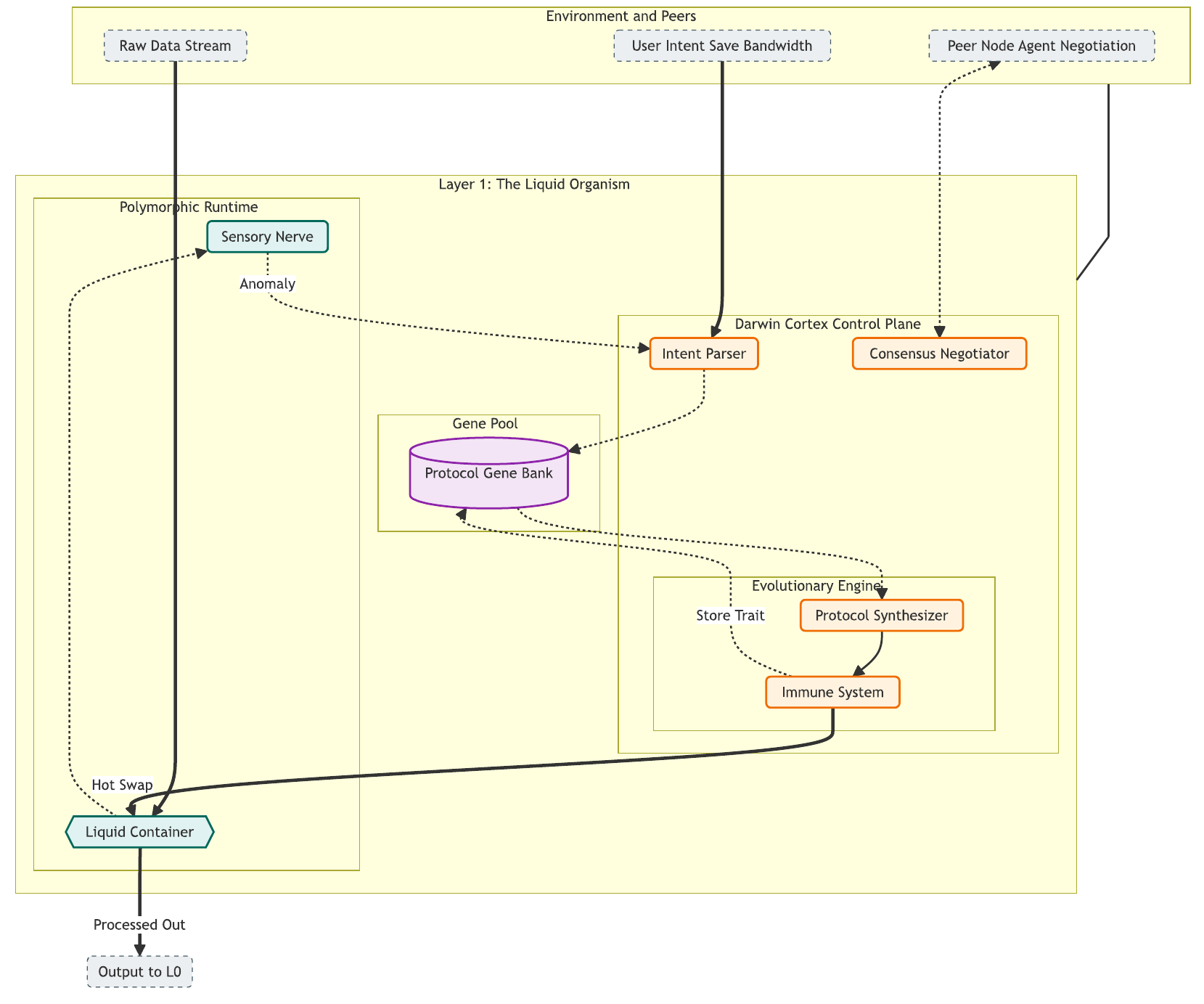}}
\caption{The Bionic Dual-Loop Feedback System of DarwinNet: A transition from rigid rules to organic adaptation.}
\label{fig:layer1}
\end{figure*}

As illustrated in Fig. \ref{fig:layer1}, the Fluid Cortex is architected as a \textit{bionic Dual-Loop feedback system}, conceptualizing the network node as a self-evolving \textit{liquid organism} \cite{rani2022optimized, weyns2023vision}. This design explicitly bifurcates the processing logic into two cognitive pathways:

\begin{itemize}
    \item \textbf{The Polymorphic Runtime (System 1):} Occupying the lower tier, this \textit{body} serves as the \textit{Fast Path}. It utilizes deformable, liquid-like containers to carry real-time data streams. Its primary objective is the pursuit of ultimate execution efficiency, operating on the evolved \textit{reflexes} (bytecode) generated by the upper layer.
    \item \textbf{The Darwin Cortex (System 2):} Occupying the upper tier, this \textit{brain} serves as the \textit{Slow Path}. It is responsible for high-level perception, identifying shifts in user intent and environmental anomalies through its \textit{sensory nerves.}
\end{itemize}
The synergy between these two loops enables a sophisticated evolutionary reflex: when the sensory nerves trigger an evolution request, the Darwin Cortex retrieves historical experiences from a \textit{Gene Pool} and synthesizes a candidate protocol via the \textit{Evolution Engine}. Crucially, before deployment, the candidate logic must pass a rigorous \textit{Immune System}—a safety verification layer that performs formal checks and sandboxed validation. Upon approval, the system executes a millisecond-level \textit{Hot Swap} on the underlying polymorphic containers. This architecture facilitates a fundamental paradigm shift from \textit{rigid rules} to \textit{organic adaptation,} ensuring that the network can emerge into optimal communication forms in response to environmental flux while maintaining strict endogenous security.

\subsubsection{Layer 2: The Darwin Cortex (The Cognitive Brain).} 
The Darwin Cortex is the intelligent decision-making center where autonomous agents and LLMs reside \cite{nascimento2023self, wang2025evoagentx}. It functions as the \textit{Slow Path} (System 2), responsible for high-level reasoning, intent sensing, and protocol synthesis \cite{wei2022chain}. By analyzing environmental pressure and user intent, Layer 2 generates the specific protocol \textit{mutations} (bytecode) that are then pushed down to Layer 1 for execution \cite{wu2024netllm, liang2024generative}.

\subsection{The Evolutionary Loop}
The evolution of communication in DarwinNet is achieved through a closed-loop feedback mechanism, as illustrated in the following four stages:

\begin{enumerate}
    \item \textbf{Sensing:} The network continuously monitors Quality of Service (QoS) metrics (e.g., packet loss, jitter) and listens for changes in business intent or environmental anomalies (e.g., novel congestion patterns or security threats).
    \item \textbf{Mutation (LLM Generation):} When a performance bottleneck or intent mismatch is detected, the Darwin Cortex triggers a \textit{directed mutation.} Unlike random biological mutations, the LLM utilizes its vast knowledge base to synthesize a specific protocol snippet—tailored to the current context—and compiles it into restricted WASM bytecode \cite{chen2021evaluating}.
    \item \textbf{Negotiation:} Before deployment, the communicating nodes must reach a consensus on the new protocol. This involves exchanging code hashes, verifying digital signatures, and performing a dry-run within the sandbox to ensure compatibility and security.
    \item \textbf{Hot Swap (Deployment):} Once consensus is reached, the Fluid Cortex performs a millisecond-level logic reload. The old protocol stack is \textit{melted} and replaced by the new, optimized bytecode, completing the evolutionary leap without packet loss or session termination.
\end{enumerate}

\subsection{Protocol Solidification Index (PSI)}
To quantify the maturity and efficiency of an evolving DarwinNet, we propose the \textit{Protocol Solidification Index (PSI)}, denoted as $M(t)$.

The PSI is defined by the following ratio:
\begin{equation}\label{eq:mature}
M(t) = 1 - \frac{N_{agent}}{N_{total}}
\end{equation}
where $N_{total}$ represents the total number of communication cycles and $N_{agent}$ is the number of cycles requiring high-level intervention from the Darwin Cortex (LLM). Furthermore, in practical implementations, $N_{total}$ can be substituted with a sliding window of size $W$ to capture the instantaneous solidification rate, thereby enhancing the sensitivity of the PSI to environmental fluctuations and transient shocks.

\textbf{Physical Significance and Maturity Measurement:} 
The PSI measures the \textit{collapse} of a system from Slow Thinking (System 2) to Fast Thinking (System 1). 
\begin{itemize}
    \item \textbf{Chaotic Phase ($M \approx 0$):} In the early stages or during environmental upheaval, the system is highly fluid; the LLM is frequently awakened to resolve conflicts and invent new adaptations, resulting in high flexibility but lower execution efficiency.
    \item \textbf{Stabilized Phase ($M \to 1$):} As the system matures, optimal interaction patterns are \textit{solidified} into efficient bytecode. The Darwin Cortex enters a dormant state, and the network operates at physical limit speeds using its evolved \textit{reflexes.}
\end{itemize}
This metric reflects the \textit{breathing} nature of DarwinNet: when a network faces a novel attack or environment shift, $M(t)$ temporarily drops as the system \textit{wakes up} to evolve, then rises again as the new solution is codified and optimized \cite{clark2003knowledge}.

\subsection{Safety Guardrails and Protocol Alignment}
To mitigate the risks associated with AI-generated logic, DarwinNet implements a multi-stage safety filter. First, the \textit{Evolution Engine} is restricted by a \textit{Policy-Aligned Template} that prevents the synthesis of known malicious communication patterns (e.g., amplification attack vectors). Second, the \textit{Immune System} performs static analysis and formal verification on the synthesized WASM bytecode before execution. Finally, the L0 anchor enforces runtime monitoring, where any protocol mutation that exceeds pre-set traffic or resource quotas is instantly melted and reverted to a known-safe baseline state. This multi-layered defense-in-depth ensures that the self-evolving nature of DarwinNet remains aligned with human intent and network security policies.

\section{Experimental Evaluation}\label{sec:experiments}
This section presents the experimental validation of DarwinNet through a discrete-event simulation framework\footnote{https://github.com/wolfbrother/DarwinNet}. Our objective is to evaluate the system’s ability to evolve from high-latency intelligent reasoning to high-efficiency native execution.

\subsection{Simulation Methodology and Fault Modeling}
To simulate \textit{protocol mismatch events} (anomalies where existing protocols fail to meet environmental or intent requirements), we adopt the \textit{Crow-AMSAA (Duane) Power Law process} \cite{duane2007learning, tang2015comparison}. This model is particularly suited for systems undergoing continuous reliability improvement through iterative \textit{discovery and repair} cycles \cite{huang2005performance}.

\begin{figure}[htbp]
\centerline{\includegraphics[width=\linewidth]{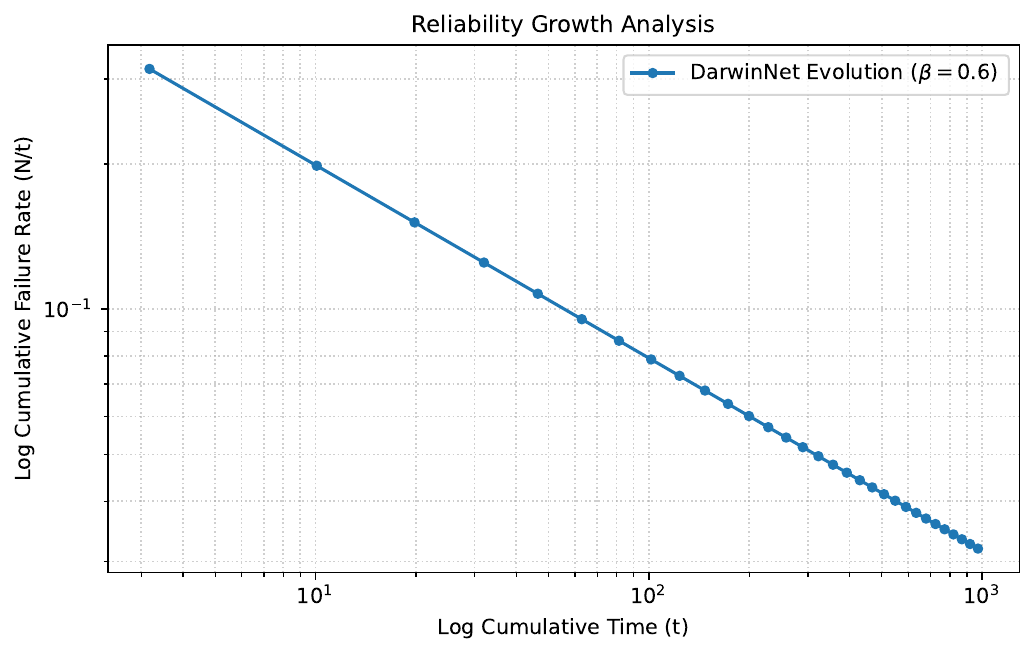}}
\caption{Reliability growth analysis using the Crow-AMSAA Power Law model, demonstrating the diminishing frequency of protocol mismatch events as the system learns and adapts.}
\label{fig:reliability}
\end{figure}

The occurrence of protocol failure events follows the intensity function defined in Eq. \ref{eq:fail}. In our simulation, we set the shape parameter $\beta = 0.6$ (where $\beta < 1$) to model the \textit{evolutionary learning effect}. As illustrated in the Duane plot in Fig. \ref{fig:reliability}, the cumulative failure rate exhibits a \textbf{linear downward trend} on a log-log scale.

Mathematically, since the cumulative number of events $N(t)$ follows $N(t) = \alpha t^\beta$, the cumulative failure rate is given by $C(t) = N(t)/t = \alpha t^{\beta-1}$. In the log-log domain, this relationship transforms into a linear equation. The constant negative slope  observed in our results provides a formal verification of the system's predictable reliability growth. This validates the hypothesis that as DarwinNet adapts to environmental stressors, the frequency of \textit{unseen} edge cases decreases over time, representing a successful transition from a chaotic state to a governed evolutionary trajectory.

We implement a dual-path state machine to map these simulation events to performance metrics:
\begin{itemize}
    \item \textbf{Slow Path (Evolutionary Intervention):} Triggered by each Crow-AMSAA failure event, this path involves LLM-based reasoning and code synthesis, characterized by high latency and computational overhead.
    \item \textbf{Fast Path (Solidified Execution):} During stable intervals, the system executes previously synthesized WASM bytecode within the Fluid Cortex ($N_{fast}$), achieving near-native performance.
\end{itemize}

\subsection{Quantitative Analysis of Protocol Solidification}
\begin{figure}[htbp]
\centerline{\includegraphics[width=\linewidth]{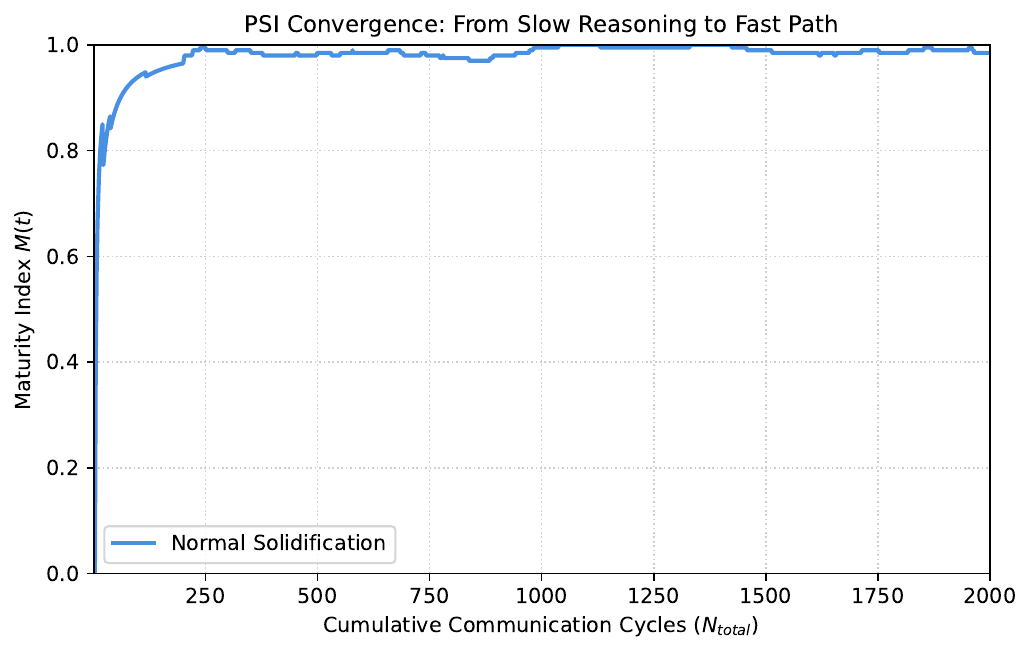}}
\caption{PSI convergence over cumulative communication cycles, illustrating the transition from agent-driven reasoning to solidified native execution.}
\label{fig:psi_convergence}
\end{figure}

The primary metric for evaluating DarwinNet's maturity is the PSI in Eq. \ref{eq:mature}. For the following quantitative analysis, the PSI is calculated using a sliding window of $W=50$ cycles to better reflect the system's instantaneous stability and responsiveness to environmental changes. Fig. \ref{fig:psi_convergence} captures the \textit{collapse efficiency} of the system. Initially, in the \textit{Chaotic Phase}, $M(t)$ remains low due to frequent agent-driven reasoning (System 2). However, as interaction patterns are successfully synthesized into executable bytecode, $M(t)$ rapidly converges toward 1.0 within the first 200 cycles. This indicates that the LLM has effectively \textit{compiled} environmental requirements into efficient binary \textit{reflexes} (System 1), allowing the system to operate at near-native performance.

\begin{figure}[htbp]
\centerline{\includegraphics[width=\linewidth]{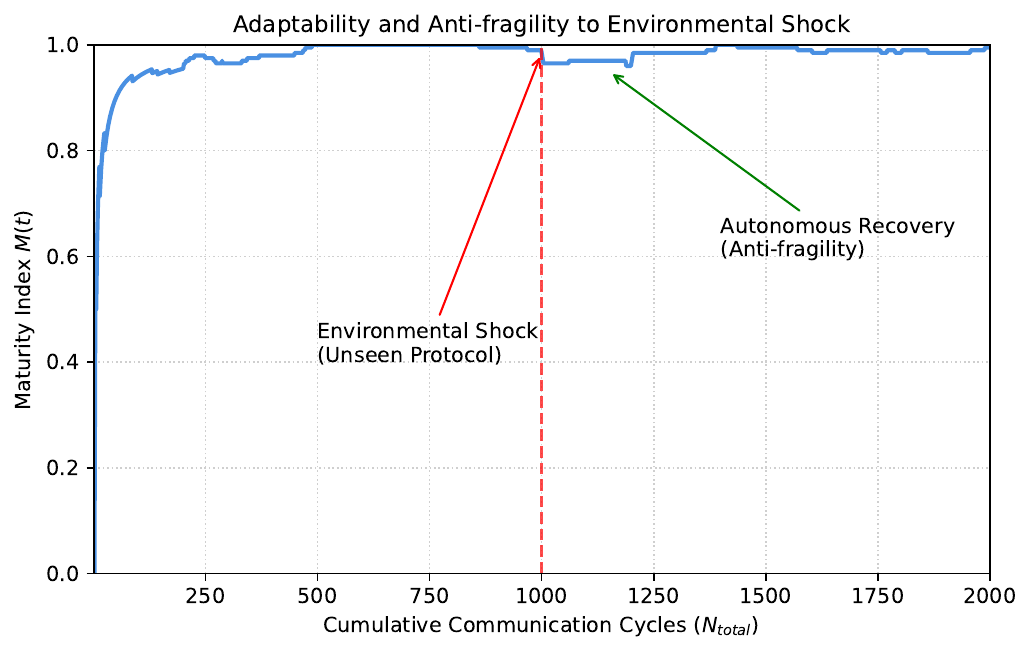}}
\caption{System adaptability and anti-fragility under environmental shock, showing a temporary dip in PSI followed by rapid autonomous recovery through protocol re-mutation.}
\label{fig:antifragility}
\end{figure}

To evaluate the system's robustness, an \textit{environmental shock} (representing an unseen, undefined protocol conflict) is introduced at $N_{total}=1000$. As illustrated in Fig. \ref{fig:antifragility}, the PSI experience a controlled dip as the Darwin Cortex is re-awakened to handle the anomaly. Crucially, the system demonstrates anti-fragility: it does not merely return to its prior state but rapidly re-solidifies, reaching a new maturity equilibrium within dozens of cycles. This \textit{V-shaped} recovery confirms that DarwinNet treats unexpected errors as fuel for further evolution rather than causes of catastrophic failure.

\subsection{Performance Gain and Engineering Feasibility}
\begin{figure}[htbp]
\centerline{\includegraphics[width=\linewidth]{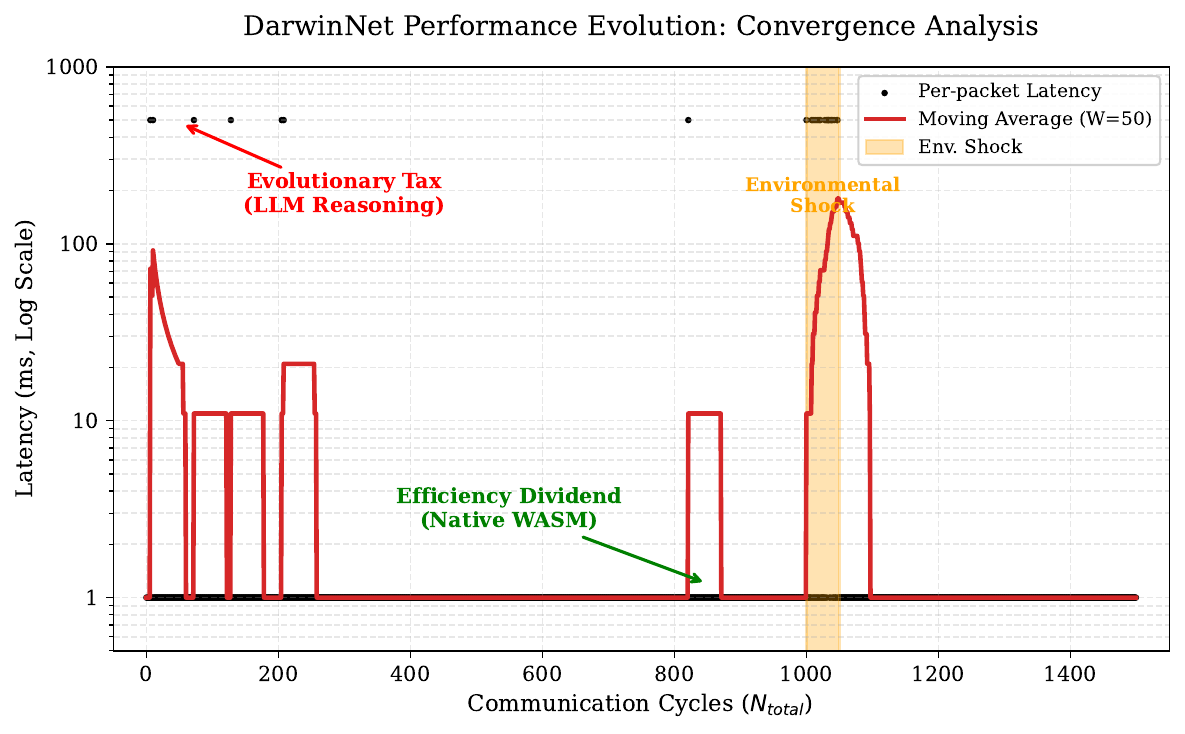}}
\caption{Latency performance evolution on a logarithmic scale, highlighting the transition from the high-latency \textit{Evolutionary Tax} (LLM-based reasoning) to the near-native \textit{Efficiency Dividend} (solidified execution).}
\label{fig:latency}
\end{figure}

The engineering viability of DarwinNet is evaluated by comparing the performance profiles across its lifecycle. Fig. \ref{fig:latency} visualizes the latency evolution on a logarithmic scale. To ensure statistical consistency with the PSI analysis, we calculate the moving average of operational latency using a sliding window of $W=50$ cycles, allowing for a synchronized observation of the correlation between protocol maturity and execution efficiency.

In the \textit{Initial Evolutionary Period}, the system pays a significant \textbf{Evolutionary Tax}, with latency spikes reaching 500ms during LLM-based protocol reconstruction. As shown by the discrete points at the top of Fig. \ref{fig:latency}, these peaks represent the effortful System 2 reasoning required to navigate unknown environmental states or complex user intents. However, as the protocol solidifies into optimized bytecode, the moving average latency drops by three orders of magnitude, eventually aligning with the 1ms physical execution limit of the WASM runtime. This transition into the \textbf{Efficiency Dividend} zone proves that the high cost of agent intervention is a transient investment that yields long-term, near-optimal operational efficiency.

To further validate the system's robustness, an \textbf{environmental shock} is introduced at $N_{total}=1000$. This event triggers a temporary re-awakening of the Darwin Cortex to address the anomaly. Crucially, the system exhibits \textbf{anti-fragility}: it does not merely recover but rapidly re-solidifies its performance, reaching a new equilibrium within a few dozen cycles. This \textit{V-shaped} recovery demonstrates that DarwinNet treats environmental stressors as catalysts for further evolution rather than causes of failure, ensuring sustained performance in hyper-heterogeneous environments.

\subsection{Convergence Verification under Crow-AMSAA Constraints}
Finally, by correlating the reliability growth in Fig. \ref{fig:reliability} with the windowed maturity index in Fig. \ref{fig:psi_convergence}, we verify the mathematical convergence of the DarwinNet framework. Although the sliding window ($W=50$) introduces local fluctuations—particularly during the re-awakening of the Darwin Cortex at cycle 1000—the global trend of the PSI consistently aligns with the reliability growth predicted by the Crow-AMSAA power law model. 

The experimental results confirm that under stable environmental constraints, the protocol reaches a \textit{Solidification Equilibrium}. At this stage, the frequency of protocol mismatch events ($\lambda(t)$) approaches zero, and the PSI stabilizes near 1.0, indicating that the protocol has sufficiently adapted to the environment to require no further agent intervention. This convergence validates that DarwinNet is not an infinitely chaotic system but a stable, commercially viable architecture capable of reaching physical performance limits through autonomous, bounded growth.

\section{Conclusion}\label{sec:conclusion}
This paper has presented \textit{DarwinNet}, a fundamental departure from the decades-old static networking paradigm toward a bio-inspired, self-evolving architecture. By shifting the core of network communication from \textit{design-time} standardization to \textit{runtime} autonomous growth, DarwinNet addresses the chronic issues of protocol ossification and structural fragility. The experimental validation, grounded in the Crow-AMSAA reliability growth model, confirms that DarwinNet is not merely a theoretical construct but a viable engineering framework. It provides a path toward Self-Defining Networks that complement the ongoing 6G standardization efforts, offering a dynamic optimization layer for the most demanding and unpredictable environments.

Beyond technical optimization, DarwinNet points toward an ultimate divergence of silicon and carbon intelligence. By decoupling protocol synthesis from the need for human interpretability, we move past the inefficiencies of human-centric design. In this new era, the network is no longer a static tool but a living organism; the role of the engineer shifts from the writer of code to the architect of evolutionary constraints, ensuring that while the machines language may become unintelligible to us, their goals remain anchored in mathematical truth.

Future work will focus on optimizing the \textit{evolutionary tax} during the initial learning phases to reduce LLM-related energy overhead and exploring the scalability of DarwinNet in hyper-heterogeneous, massive-scale IoT environments. As artificial intelligence continues to permeate every layer of the computing stack, the transition from \textit{designed protocols} to \textit{grown protocols} will be a cornerstone of the next generation of resilient, self-optimizing infrastructure.

\section*{Acknowledgment}
The authors contributed collectively to the research. However, the specific strategic visions and conceptual interpretations presented herein are attributed to the corresponding author (xujinliang@caict.ac.cn, jlxufly@gmail.com) to reflect his current stage of thinking, and do not necessarily represent the official views of the other authors or the affiliated institutions.

\bibliographystyle{ieeetr}
\bibliography{Ref}

\end{document}